\pdfoutput=1

\documentclass[11pt]{article}

\usepackage[final]{acl}

\usepackage{amsmath,amsfonts,bm}

\def\eqref#1{equation~\ref{#1}}

\def\1{\bm{1}}

\DeclareMathAlphabet{\mathsfit}{\encodingdefault}{\sfdefault}{m}{sl}
\SetMathAlphabet{\mathsfit}{bold}{\encodingdefault}{\sfdefault}{bx}{n}

\usepackage{hyperref}
\usepackage{url}
\usepackage{booktabs}
\usepackage{graphicx}
\usepackage{xcolor}
\usepackage{multicol}
\usepackage{multirow}
\usepackage{times}
\usepackage{xspace}
\usepackage{amssymb}
\usepackage{pifont}
\newcommand{\cmark}{\ding{51}}%
\newcommand{\xmark}{\ding{55}}%

\definecolor{ggreen}{HTML}{00A64F}
\definecolor{light-gray}{gray}{0.9}

\newcommand*{\tightcolorbox}[2]{%
    \begingroup\setlength{\fboxsep}{1pt}%
        \colorbox{#1}{{\hspace*{2pt}\vphantom{Ay}#2\hspace*{2pt}}}%
    \endgroup
}
\newcommand*{\code}[1]{\tightcolorbox{light-gray}{\texttt{#1}}}
\newcommand*{\modelname}[1]{{\textsc{#1}}}
\newcommand*{\datasetname}[1]{{\textsc{#1}}}

\title{The LLM Language Network:\\A Neuroscientific Approach for Identifying Causally Task-Relevant Units}

\author{
    Badr AlKhamissi$^1$ \quad 
    Greta Tuckute$^2$ \quad 
    Antoine Bosselut\thanks{Equal Supervision}$^{,1}$ \quad 
    Martin Schrimpf$^{*,1}$ \\
    $^1$EPFL \quad $^2$MIT 
}

\begin{document}

\maketitle

\begin{abstract}

Large language models (LLMs) exhibit remarkable capabilities on not just language tasks, but also various tasks that are not linguistic in nature, such as logical reasoning and social inference.
In the human brain, neuroscience has identified a \emph{core language system} that selectively and causally supports language processing.
We here ask whether similar specialization for language emerges in LLMs.
We identify language-selective units within 18 popular LLMs, using the same localization approach that is used in neuroscience.
We then establish the causal role of these units by demonstrating that ablating LLM language-selective units -- but not random units -- leads to drastic deficits in language tasks.
Correspondingly, language-selective LLM units are more aligned to brain recordings from the human language system than random units.
Finally, we investigate whether our localization method extends to other cognitive domains: while we find specialized networks in some LLMs for reasoning and social capabilities, there are substantial differences among models.
These findings provide functional and causal evidence for specialization in large language models, and highlight parallels with the functional organization in the brain.\footnote{Code available via \href{https://github.com/bkhmsi/llm-localization}{github.com/bkhmsi/llm-localization}} %

\end{abstract}

\begin{figure}[th!]
    \centering
    \includegraphics[width=1\linewidth]{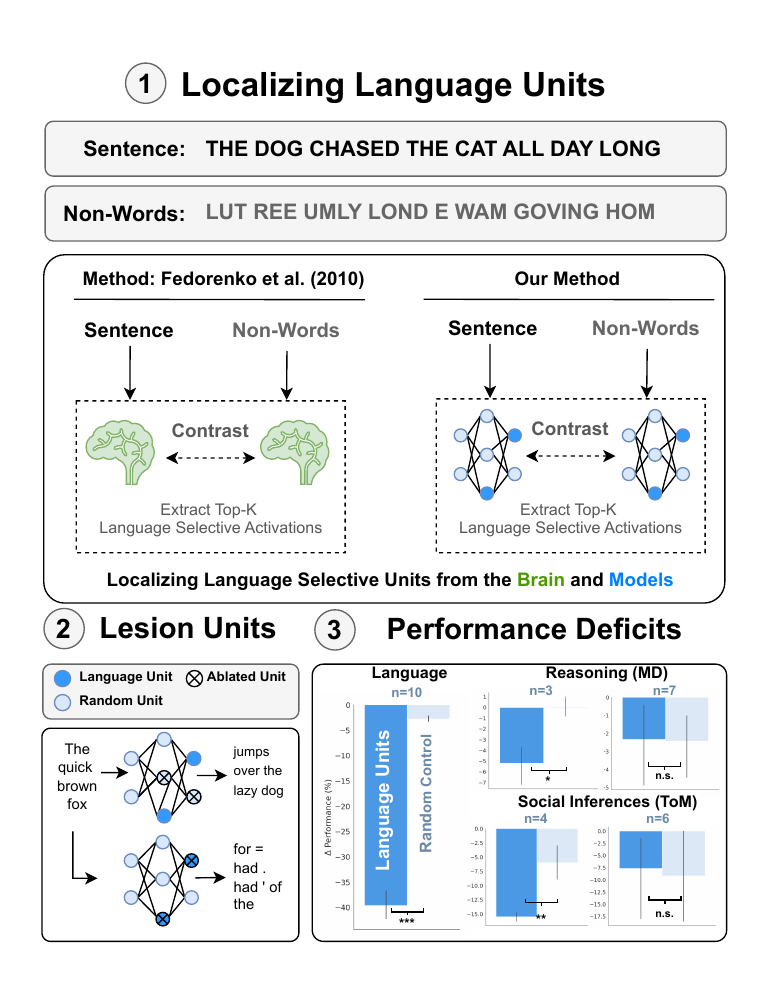}
    \caption{
    \textbf{Identifying Specialized and Causally Task-Relevant Units in LLMs.}
    \textbf{(1)} To identify language-selective units, we compare unit activations in response to language (sentences) versus a matched control condition (lists of non-words), and identify the units that exhibit the strongest selectivity to sentences over non-words. The same method is used in neuroscience to localize the human brain's language network (e.g., \citealp{Fedorenko2010NewMF}).
    \textbf{(2)} Testing the causal role of the identified language-selective units, we ablate those units as well as a set of random units, and 
    \textbf{(3)} compare the resulting performance drop. Ablating 1\% of LLM language units leads to vast language deficits ($p<5^{-238}$) for all models tested. 
    Beyond language, only a few models exhibit specialization for reasoning (n=3, $p<5^{-2}$, Multiple Demand network) and social inferences (n=4, $p<5^{-5}$, Theory of Mind network). 
    Plots averaged across \textit{n} LLMs each; random control repeated with 3 different seeds.
    }
    \label{fig:localization-method}
\end{figure}

\section{Introduction}


\begin{table*}[]
\centering
\begin{tabular}{@{}lll@{}}
\toprule
\textbf{Model}                 & \textbf{Ablate Language Units}      & \textbf{Ablate Random Units}     \\ \midrule
\textbf{Gemma-2B}              & 11 liquido \_ sota(.)uggoon3 & jumped over the lazy lamb. \\
\textbf{Phi-3.5-Mini-Instruct} & AME.AME and:ough.. MAR       & jumps over the lazy dog.   \\
\textbf{Falcon-7b}             & SomeSReadWhenISearchSome     & jumps over the lazy dog.   \\
\textbf{Mistral-7B-v0.3}       & foxfool foolfoolfoolfool     & jumps over the lazy dog.   \\ 
\textbf{LLaMA-3.1-8B-Instruct} & \_ of\_An\_O\_of\_An\_O\_of  & jumps over the lazy dog.   \\ \bottomrule
\end{tabular}
\caption{
   \textbf{Disruption of Language Modeling Abilities}
   Continuations of the prompt ``The quick brown fox'' across five different models, following the ablation of the top 1\% of language-selective units compared to the ablation of an equivalent number of randomly selected units. The baseline generation without ablation for all models was ``jumps over the lazy dog.''
}
\label{tab:lesion-gens}
\end{table*}

Recent advancements in large language models (LLMs) have revealed their potential to perform far more than language processing tasks, showcasing abilities in reasoning \cite{sun2023surveyreasoning}, problem-solving \cite{giadikiaroglou2024puzzle}, and even mimicking aspects of human Theory of Mind \cite{street2024llmstom}. Despite these impressive feats, the internal workings of LLMs remain poorly understood, especially in relation to how specific components of these models contribute to manifesting distinct cognitive functions. 

The field of neuroscience has made significant strides in mapping out the functional organization of the human brain, for instance by identifying specialized cognitive networks such as the language network \cite{Fedorenko2010NewMF,Fedorenko2024}, the Multiple Demand network \cite{duncan2010multiple,assem2020domain}, and the Theory of Mind network \cite{saxe2013people}, each underlying distinct cognitive behaviors. 
In this paper, we draw inspiration from neuroscience to investigate whether similar functional specialization exists within LLMs. 

\begin{figure*}
    \centering
    \includegraphics[width=1\linewidth]{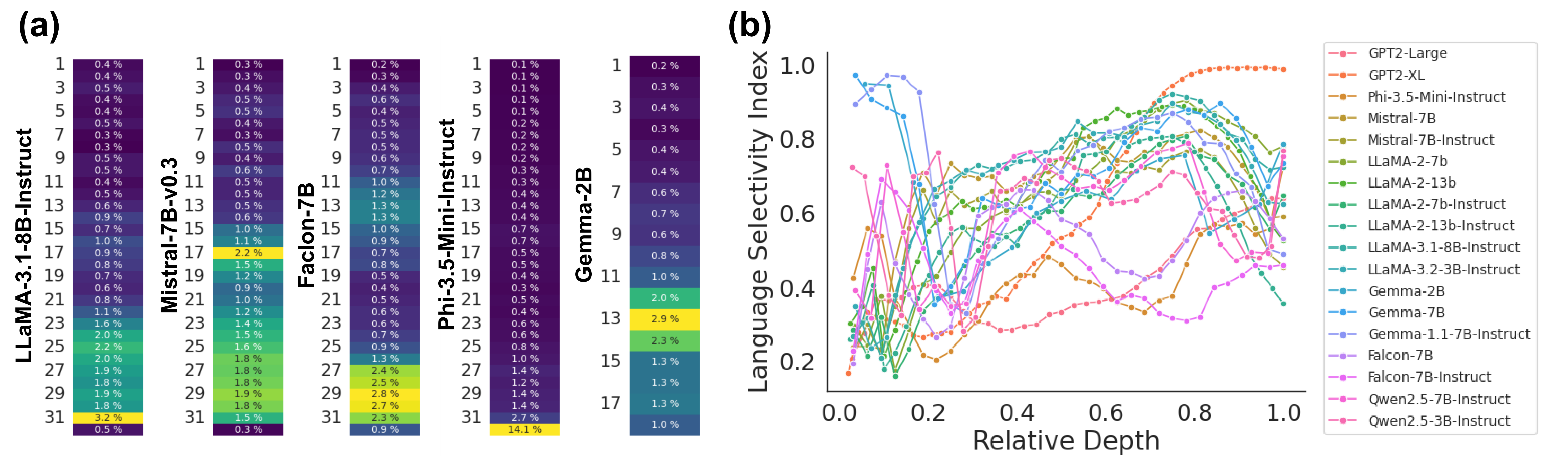}
    \caption{
        \textbf{Distribution of Language Units Across Layers}.
         \textbf{(a)} The distribution of the top 1\% most language-selective units across layers in a sample of five different models. The models are displayed from top to bottom, with each layer labeled by the percentage of units identified as belonging to the top 1\% language-selective units.
        \textbf{(b)} The language selectivity index for all models in the study (n=18) plotted against the relative depth of the layers.
    }
    \label{fig:lang-map}
\end{figure*}

Specifically, we use the same localizer experiments developed by neuroscientists to identify functional brain regions. These experiments contrast activations between target conditions of interest (e.g., sentences) and perceptually matched control conditions (see Section \ref{sec:localizer}). We discover that, much like the human brain, there exists a set of units in LLMs that are critical for language processing, analogous to the human language network \cite[][Fig. \ref{fig:lang-map}]{Fedorenko2024}. We find that these units show similar response patterns as those observed in the human language areas \cite{Shain2024a,schrimpf-pnas}, and, moreover, demonstrate selectivity for natural language compared to mathematical equations and computer code, much like the human brain \cite{ivanova_comprehension_2020,fedorenko2011functional,Fedorenko2024}. 

Further, ablating even a small percentage of these language-selective units results in a significant decline in language performance, demonstrated qualitatively in Table \ref{tab:lesion-gens} and quantitatively in Figure \ref{fig:lesion-study} through benchmarks like SyntaxGym \cite{gauthier-etal-2020-syntaxgym}, BLiMP \cite{Warstadt2019BLiMPTB}, and GLUE \cite{wang-etal-2018-glue}. 
Finally, the language-selective units show stronger alignment with the brain's language network compared to randomly sampled units, especially when selecting a small number of units to predict brain activity (Figs. \ref{fig:response-profiles}, \ref{fig:brainscore}). 
Despite substantial evidence for the existence of a language network in all LLMs we tested, we only found evidence of units selective for social (Theory of Mind) and reasoning (Multiple Demand) tasks in a subset of models (Figure \ref{fig:md-lesion}).


\section{Preliminaries}



\paragraph{The Human Language Network.}
The human language network comprises a set of brain regions that are functionally defined by their increased activity to language inputs over perceptually matched controls (e.g., lists of non-words) \citep[][Section \ref{sec:localizer}]{Fedorenko2010NewMF,Lipkin2022}. These regions are predominantly localized in the left hemisphere, within frontal and temporal areas, and demonstrate a strong selectivity for language processing over various non-linguistic tasks such as music perception \citep{Fedorenko2012music, chen2023human} and arithmetic computation \citep{fedorenko2011functional, monti2012thought}. Crucially, these regions exhibit only weak activation in response to meaningless non-word stimuli, whether during comprehension or production \citep{Fedorenko2010NewMF, hu2023precision}. 
This high degree of selectivity is well-established through neuroimaging studies and is further supported by behavioral data from aphasia studies: In individuals with damage confined to these language areas, linguistic abilities are significantly impaired, while other cognitive functions—such as arithmetic computations \citep{Benn2013, Varley2005}, general reasoning \citep{Varley2000}, and Theory of Mind \citep{Siegal2006}—remain largely intact. In addition to language-specific systems, the brain supports higher-level cognition through distinct networks that handle demanding tasks and social reasoning.

\paragraph{The Multiple Demand Network.}
The Multiple Demand Network (MD), encompassing bilateral frontal, parietal, and temporal regions, is activated during cognitively demanding tasks, showing a consistent ``hard $>$ easy'' response across various task types (e.g., spatial, verbal, mathematical; \citealp{Duncan2000, Fedorenko2013, Shashidhara2020}). This network underpins key cognitive functions such as working memory, cognitive control, and attention, and is linked to fluid intelligence \cite{Woolgar2010, Assem2020}. 

\paragraph{The Theory of Mind Network.} The Theory of Mind (ToM) network, primarily located in the bilateral temporo-parietal junction and cortical midline, is involved in reasoning about mental states—whether one's own or others' \cite{Saxe2003, Gallagher2000, Saxe2006}. Functionally and anatomically distinct from the language network, the ToM network is engaged across different content types (e.g., verbal, non-verbal) and is engaged in understanding non-literal language such as sarcasm, and for discourse comprehension where multiple perspectives need to be inferred \cite{KosterHale2013, Hauptman2023}.


\section{Localizing the Language Network}
\label{sec:localizer}

The human language network is defined \emph{functionally} rather than anatomically which means that units are chosen according to a `localizer' experiment \citep{saxe2006divide}. Specifically, the language network is the set of neural units (e.g., voxels/electrodes) that are more selective to sentences over a perceptually-matched control condition (e.g., lists of nonwords) \citep{Fedorenko2010NewMF} as illustrated in Figure \ref{fig:localization-method}.
In previous studies comparing artificial models to brain activity in the language network, units were selected by evaluating representations at different model layers and choosing the ones that maximized brain alignment \citep{schrimpf-pnas,goldstein_shared_2022,caucheteux2022brains,tuckute2024driving}. However, LLMs learn diverse concepts and behaviors during their considerable pretraining, not all of which are necessarily related to language processing.
Therefore, we here characterize the language units in LLMs using functional localization as is already standard in neuroscience. This approach comes with the advantage of comparability across different models since we can choose a fixed set of units which are localized independently of the critical experiment or modality. 


Specifically, we present each LLM with 240 unique 12-word-long sentences and 240 unique strings of 12 non-words used in neuroscience \cite{Fedorenko2010NewMF}, ensuring matched sequence lengths across conditions. Human participants are typically exposed to 96 sentences/non-word strings during an experimental fMRI session \cite{Lipkin2022}. 
We then capture the activations from the units at the output of each Transformer block for each stimulus. We define the model's language network as the top-k units that maximize the difference in activation \emph{magnitude} between sentences and strings of non-words, measured by positive t-values from a Welch's t-test. This localization method selects a targeted set of units across the entire network, rather than restricting the representations to a single a priori stage as done in prior work \cite{schrimpf-pnas,goldstein_shared_2022,caucheteux2022brains,tuckute2024driving}. We examine unit activations after each Transformer block. For instance, for a model like \modelname{LLaMA-3-8B} \citep{llama3} which consists of 32 Transformer blocks and a hidden dimension of 4096, we consider $32 \times 4096 = 131,072$ units, from which we select the most language selective units as the model's ``language network''.


\section{Experimental Setup}

\paragraph{Models}
We selected 10 autoregressive LLMs from a diverse range of model families to investigate their language-selective units: \modelname{GPT2-\{Large, XL\}} \cite{gpt2}, \modelname{LLaMA-2-\{7B, 13B\}} \cite{llama2}, \modelname{LLaMA-3.1-8B-Instruct} \cite{llama3}, \modelname{Mistral-7B-v0.3} \cite{jiang2023mistral}, \modelname{Falcon-7B} \cite{almazrouei2023falcon}, \modelname{Phi-3.5-Mini-Instruct} \cite{abdin2024phi}, and \modelname{Gemma-\{2B, 7B\}} \cite{team2024gemma}. The models were downloaded from the HuggingFace platform \cite{Wolf2019HuggingFacesTS}.

\paragraph{Language Benchmarks}
To assess the significance of the localized units on the models' linguistic abilities, we utilize three widely used benchmarks that measure formal linguistic competence \cite{mahowald2024dissociating}. First, SyntaxGym \cite{gauthier-etal-2020-syntaxgym} offers 30 subtasks focused on evaluating syntactic knowledge. Second, BLiMP \cite{Warstadt2019BLiMPTB} contains 67 subtasks, each consisting of 1,000 minimal pairs designed to test contrasts in syntax, morphology, and semantics. Third, GLUE \cite{wang-etal-2018-glue} includes 8 subtasks aimed at assessing the models' broader language understanding.
The models are evaluated by calculating the negative log-likelihood of each candidate answer given the context, selecting the candidate that minimizes surprisal as the model's prediction. This method, commonly used in psycholinguistics, has been shown to correlate with human behavioral measures \cite{Smith2013}. SyntaxGym and BLiMP are evaluated in a zero-shot setting, while GLUE tasks are tested with 2-shot examples in context to achieve reasonable performance in the non-ablation setting.

\paragraph{Brain Alignment Benchmarks}
To validate the model language units' alignment to the human language network, we employ two approaches: i) investigating whether the model units can replicate landmark neuroscience studies that qualitatively describe the response profiles observed in the human language regions, and ii) quantitatively testing the alignment of language units with brain responses from the human language network.
For the first approach, we closely follow the analyses in \citet{Fedorenko2010NewMF} and \citet{Shain2024a}, using the same set of experimental conditions which are commonly used in neuroimaging studies examining lexical and syntactic processing. For the second approach, we measure how well the language units can predict brain activity in the human language network. To do so, we utilize the \datasetname{Tuckute2024} \cite{tuckute2024driving} and \datasetname{Pereira2018} \cite{pereira_toward_2018} benchmarks on the Brain-Score platform \cite{schrimpf_brain-score_2018,schrimpf2020}. \datasetname{Tuckute2024} consists of brain responses from 5 participants who each read 1,000 linguistically diverse sentences, while \datasetname{Pereira2018} consists of 15 subjects reading short passages presented one sentence at a time spanning various different topics. See Appendix \ref{app:brain-alignment} for more details about the datasets.

\begin{figure*}
    \centering
    \includegraphics[width=1\linewidth]{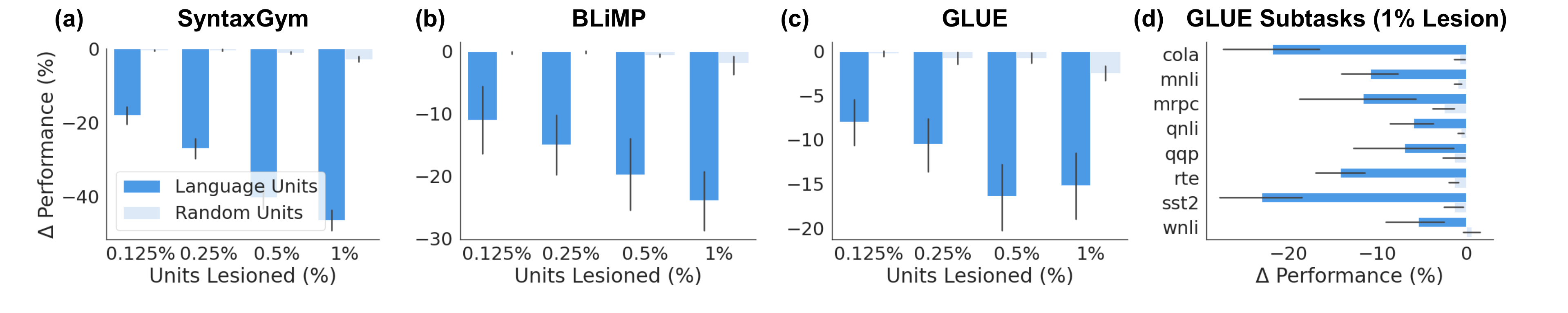}
    \caption{
        \textbf{Lesion Studies}. The average performance change after ablating the top x\% of language-selective units, compared to ablating three random sets of units for each model. Performance is evaluated across 10 models and three language benchmarks: \textbf{(a)} SyntaxGym, \textbf{(b)} BLiMP, and \textbf{(c)} GLUE, with \textbf{(d)} presenting results for individual subtasks within GLUE when ablating the top 1\% of language units.
    }
    \label{fig:lesion-study}
\end{figure*}


\section{A Specialized Language Network in LLMs}

Figure \ref{fig:lang-map}(a) shows the percentage of language units in each layer that belong to the top 1\% of the most selective units for five models analyzed in this study (additional heatmaps for other models can be found in Appendix \ref{app:language-units-loc}). Figure \ref{fig:lang-map}(b) demonstrates a language selectivity index against the relative depth of each layer across all models tested. This index is calculated by summing \(1 - p\)-values for each unit where \(p < 0.05\) after false discovery correction, and normalizing by the hidden dimension size. 
\section{The LLM Language Network is Causally Involved in Language Processing}

Table \ref{tab:lesion-gens} qualitatively illustrates the disruption in language modeling abilities when 1\% of language-selective units are ablated, in contrast to no disruption when an equivalent set of randomly sampled units is ablated. To quantify this effect, Figure \ref{fig:lesion-study} shows the average change in performance across the 10 LLMs after ablating the top-\{0.125, 0.25, 0.5, 1\}\% of language-selective units for a set of three language benchmarks which measure formal linguistic competencies \cite{mahowald2024dissociating}.  For comparison, we also measure performance changes after ablating an equivalent number of units randomly sampled from the remaining units in the model (e.g., if 0.125\% of the most language-selective units are ablated, the random units are sampled from the remaining 99.875\%), some of which may also have significant language selectivity. Random sampling results are averaged over three seeds for each model.
The results underscore the distinct role of language-selective units: ablating just 0.125\% of these units leads to a notable performance drop across all three benchmarks (Cohen's d = 0.8, large effect size; $p < 5^{-43}$). In contrast, ablating the same number of randomly sampled units has minimal impact on performance (Cohen's d = 0.1, small effect size; $p = 2^{-4}$), highlighting the unique contribution of language-selective units to the model’s linguistic capabilities.
We found that not all tasks are impacted equally (Figure \ref{fig:lesion-study}(d)): within GLUE, linguistic acceptability (COLA) and sentiment analysis (SST2) experience much more drastic performance deficits compared to Question NLI (QNLI) and Winograd NLI (WNLI). This variation may be attributable to the reliance on other non-language-specific units. We report the fine-grained results per model in Appendix \ref{app:finegrained-results}.

\section{Similarity Between the Language Network in LLMs and Brains}
\label{sec:llm-brain-similarity}

\begin{figure*}[t]
    \centering
    \includegraphics[width=1\linewidth]{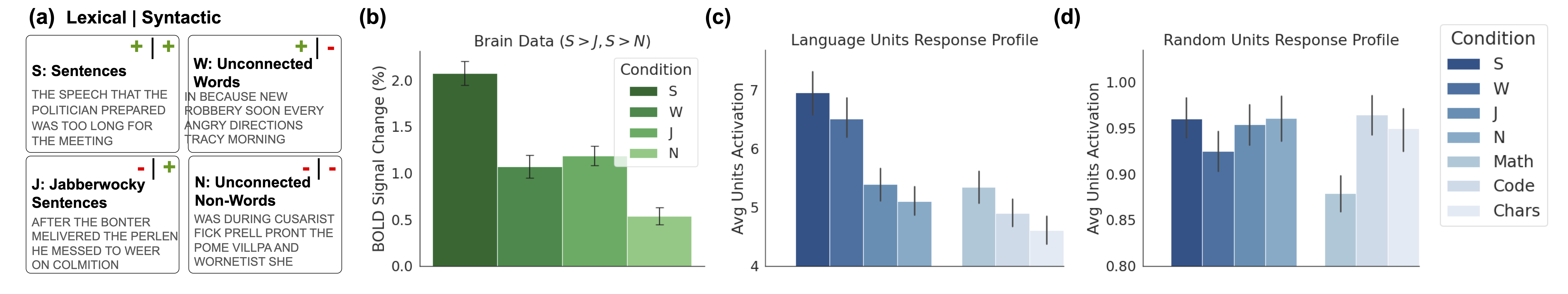}
    \caption{
    \textbf{Language-Selective Model Units Are Selective for Language and Exhibit Similar Response Profiles as the Language Network in the Brain}. 
    Brain ({\color{ggreen} green}) and model ({\color{blue} blue}) responses for Univariate Condition-Level Responses.
    \textbf{(a)} Examples of the four experimental conditions used in this analyses with the `+/-' signs denoting whether the condition contains lexical or syntactic information, respectively. 
    \textbf{(b)} Human language network responses to the four conditions; data from \citep{Shain2024a}. Brain activity is strongest to \texttt{S}, followed by \texttt{W} and \texttt{J}, and weakest to \texttt{N}.
    \textbf{(c)} Language-selective unit responses to the four conditions averaged across 10 models and condition samples.
    \textbf{(d)} Control responses from random units averaged across condition samples and 10 models, with 3 random seeds each. 
    }
    \label{fig:response-profiles}
\end{figure*}

\paragraph{Qualitatively Similar Response Profiles Between the Language Network in LLMs and Brains.}
\label{qualitative}

In this analysis, we record the responses of the localized units to the exact stimuli from four experimental conditions used in neuroscientific studies \cite{Fedorenko2010NewMF,Shain2024a}, along with a set of non-linguistic stimuli such as arithmetic equations and code. This allows us to assess how well the selectivity of localized language units generalizes to new stimuli from the same conditions (sentences and strings of non-words) and how well they map onto results from neuroscience \cite{Amalric2019, ivanova_comprehension_2020,Fedorenko2024}. 
Stimuli are presented in four conditions (examples in Figure \ref{fig:response-profiles}a): 
\code{Sentences}, denoted as \emph{S}, are well-formed sentences containing both lexical and syntactic information. 
\code{Unconnected Words}, \emph{W}, are scrambled sentences containing lexical but not syntactic information. 
\code{Jabberwocky Sentences}, \emph{J}, where content words are replaced by pronounceable non-words (such as ``pront'', or ``blay''), thus containing syntactic but not lexical information. 
\code{Unconnected Non-Words}, \emph{N}, which are scrambled Jabberwocky sentences containing neither lexical nor syntactic information. 
Note that we use a disjoint set of \code{Sentences} and \code{Non-Words}for the original functional localization (Section \ref{sec:localizer}).
The brain's language regions are highly sensitive to linguistic structure: responses to \emph{S} are numerically higher than all other conditions \cite{Fedorenko2010NewMF,Shain2024a,Fedorenko2024}. 


\begin{figure*}[ht]
    \centering
    \includegraphics[width=1\linewidth]{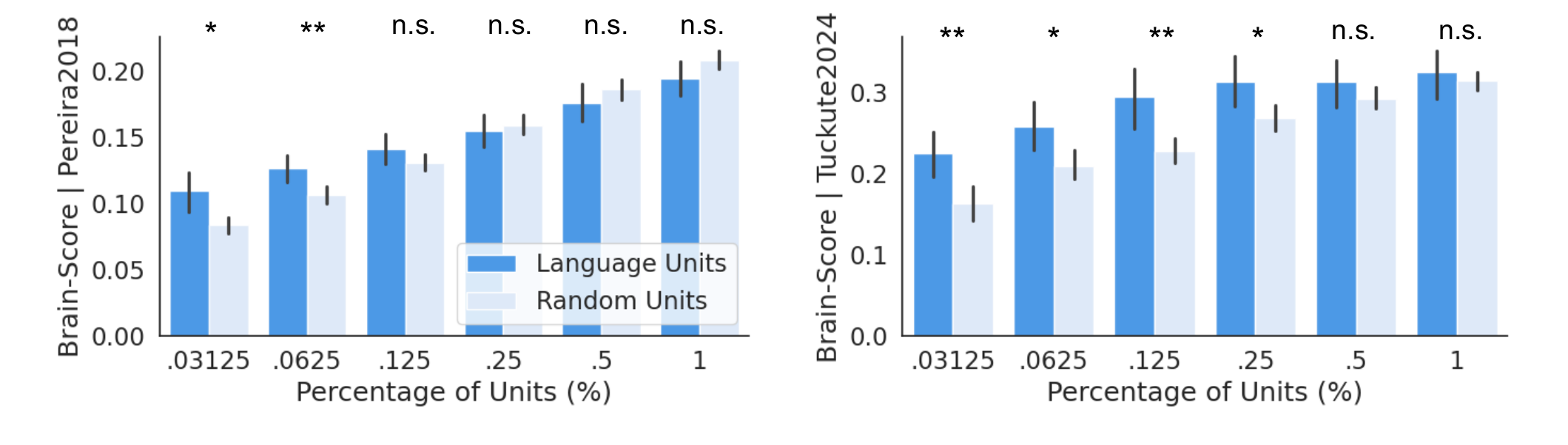}
    \caption{
        \textbf{Language Units are Aligned to Brain Data.}
        Raw Pearson correlation between predicted brain activity from the x\% of model units and actual brain activity in the human language network across 10 models. The alignment of language-selective units shows significantly greater correlation compared to the average of three sets of randomly selected units when selecting a small subset of units. Error bars represent 95\% confidence intervals calculated across models. See Table \ref{tab:num-units} for the number of units corresponding to each percentage level per model.
    }
    \label{fig:brainscore}
\end{figure*}

The LLM language units exhibit a similar response pattern to that of the brain's language network (Figure \ref{fig:response-profiles}c, first 4 bars). 
Consistent with human neuroscience \cite{fedorenko2011functional, Amalric2019, ivanova_comprehension_2020}, LLM language units are particularly selective for natural language compared to arithmetic equations, C++ code, or random sequences of characters (all matching the number of tokens and samples in the other conditions).
In contrast, responses from three sets of randomly sampled units show a different response profile (Figure \ref{fig:response-profiles}d), demonstrating that the language response profile is not ubiquitously present throughout the LLMs. 


\begin{figure*}
    \centering
    \includegraphics[width=1\linewidth]{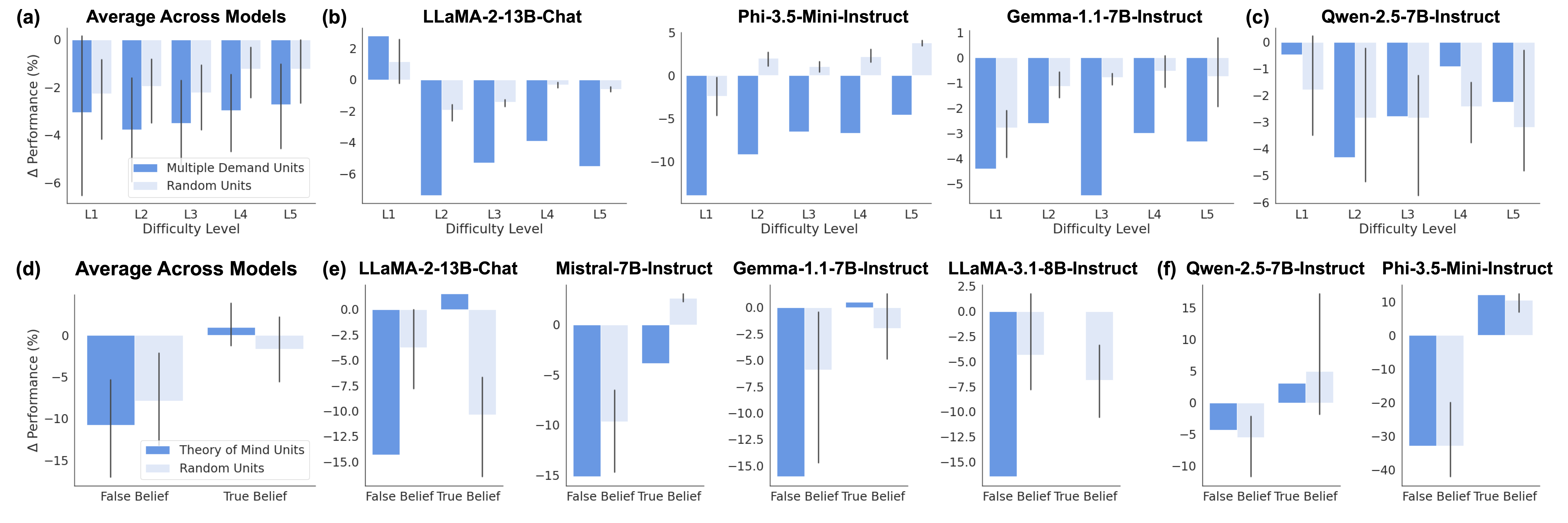}
    \caption{
        \textbf{Multiple Demand and Theory of Mind lesion study.}
        Change in performance on the (top) \datasetname{MATH} multiple-choice benchmark as a function of the difficulty level, and the (bottom) \datasetname{ToMi} multiple-choice benchmark, categorized by whether the question involves a false-belief or true-belief scenario. Results are shown after ablating the top 1\% of MD-selective and ToM-selective units respectively as well as an equivalent number of random units (across 3 seeds). \textbf{(a,d)} Average performance change for \datasetname{MATH}/\datasetname{ToMi}, across all 10 models. \textbf{(b,e)} Models where ablating MD/ToM units leads to a greater performance drop on difficult/false-belief problems compared to random unit ablation. \textbf{(c,f)} Sample of models showing no difference between ablating MD/ToM units and random units.
    }
    \label{fig:md-lesion}
\end{figure*}

\paragraph{Quantitative Alignment Between the Language Network in LLMs and Brains.}

Beyond \emph{qualitative} alignment between LLM language units and brains, we investigate the \emph{quantitative} alignment to brain data. 
Following standard practice in measuring brain alignment, we train a ridge regression to predict brain activity from model representations, using the same input stimuli presented to human participants in neuroimaging studies \cite{naselaris2011encoding,schrimpf-pnas}. We then measure the Pearson correlation between the predicted brain activations and the actual brain activations of human participants on a held-out set. This process is repeated over 10 cross-validation splits, and we report the average (mean) Pearson correlation as our final result which we here refer to as Brain-Score \cite{schrimpf_brain-score_2018,schrimpf2020}. 
Figure \ref{fig:brainscore} shows the average raw correlation when using \{0.03125, 0.0625, 0.125, 0.25, 0.5, 1\}\% of model units to predict brain activity for two neural datasets \cite{pereira_toward_2018, tuckute2024driving}. This analysis is repeated for the most language selective units, and for three sets of randomly sampled units for each of the 10 models. See Appendix \ref{app:stats-test} for more statistical tests and Appendix \ref{app:brain-alignment} for more dataset details.

\section{Localizing the Multiple Demand \& Theory of Mind Networks}

The results so far suggest that the functional localization methods used in neuroscience to identify the brain’s language network also applies effectively to LLMs. This raises a natural question: can we use the same localizers designed to identify other brain networks, such as the Theory of Mind network or the Multiple Demand network, to discover analogous networks in LLMs?

\subsection{Functional Localizers}
\label{sec:func-localizer-tom-md}

\paragraph{Multiple Demand Network}
To localize the Multiple Demand (MD) network, neuroscientists typically use either spatial or arithmetic tasks that compare brain activations during a cognitively demanding problem (a ``hard'' task) with those during an easier one \cite{Fedorenko2013}. In this work, we adapted the arithmetic MD localizer into a verbal format to explore whether a similar network can be identified in LLMs. Instead of using just the representation of the final token (as was done for localizing the language network), we average the activations across all tokens in the context before comparing the two stimulus sets. More details about the localizer stimuli can be found in Appendix \ref{app:md-localizer}.  

\paragraph{Theory of Mind Network} 
\citet{dodell2011tom} developed an efficient localizer to identify brain regions involved in Theory of Mind (ToM) and social cognition in individual participants. This was achieved by contrasting brain activation during false-belief stories---where characters hold incorrect beliefs about the world---with activation during false-photograph stories, where a photograph, map, or sign depicts an outdated or misleading world state. The false-photograph stories are verbalized to match the presentation style of the false-belief stories for consistency in the experiment. Each stimuli set consists of only 10 samples, which are sufficient to robustly identify the ToM network in the brain. Similar to the MD localizer, we average the activations across all tokens in the context before comparing the two stimulus sets. See Appendix \ref{app:tom-localizer} for more details.

\subsection{Benchmarks}

\paragraph{MATH.} The Multiple Demand (MD) network is involved during cognitively demanding tasks such as mathematical reasoning. Therefore, to evaluate the effectiveness of the MD localizer, we use the multiple choice version of the \datasetname{MATH} benchmark \cite{hendrycksmath2021} introduced by \citet{zhang2024multiplechoice}. It consists of math questions encompassing several topics ranging from ``Counting \& Probability'' to ``Geometry'' and ``Algebra''. There are 4,914 questions categorized into 5 levels of difficulty, and each one contains 4 candidate answers presented to the model. 

\paragraph{ToMi.}
To evaluate the Theory of Mind (ToM) abilities of the model, we used the \datasetname{ToMi} QA dataset preprocessed by \citet{sap2022neuralToM}, focusing only on questions that require first-order ToM reasoning. The dataset consists of 620 stories generated through a stochastic rule-based algorithm, which involves selecting two participants, an object of interest, and a set of locations or containers. These elements are combined into a narrative where the object is moved, and questions are asked about either the object's original location or its final location \cite{le-etal-2019-revisiting}. The questions include both false-belief scenarios, where a participant was absent when the object was moved, and true-belief scenarios, where the participant was present. The task requires the model to infer the ``mental states'' of the characters and the realities of the situation in the story. Each sample presents the model with an instruction, the story, the question, and two possible answers. The model's response is the answer that minimizes surprisal, measured by the negative log-likelihood.

\subsection{Models}
Given the complexity of the benchmarks used to evaluate higher-level cognitive networks, which require advanced reasoning abilities and models that are capable of following instructions for zero-shot evaluation, we transitioned all models to instruction-tuned versions.
Additionally, we included \modelname{Qwen2.5-\{3B, 7B\}-Instruct} and \modelname{LLaMA-3.2-3B-Instruct} to maintain a consistent sample size of 10 models, matching those used in the language evaluations.


\subsection{Results}
We repeat the lesion analysis performed on the language network for the Theory of Mind (ToM) and Multiple Demand (MD) selective units (top 1\%). After identifying units with the functional localizers discussed in Section \ref{sec:func-localizer-tom-md}, we measure the change in performance following the ablation of the most selective units.

\paragraph{Multiple Demand.}
Figure \ref{fig:md-lesion}(a-c) illustrates the change in performance on the \datasetname{MATH} multiple-choice benchmark for a sample of models, broken down by difficulty level. For a specialized LLM Multiple Demand network, we would expect a greater performance drop as the question difficulty increases, reflecting a more ``cognitively demanding'' task. This pattern is evident in \modelname{LLaMA-2-13B-Chat}, \modelname{Gemma-1.1-7B-Instruct}, and \modelname{Phi-3.5-Mini-Instruct}, but is less pronounced in other models. See Appendix \ref{app:finegrained-results} for results on all models.

\paragraph{Theory of Mind.}
Similar to the MD results, ToM findings are incosistent across models. Figure \ref{fig:md-lesion}(d-f) shows the results on a sample of models on the \datasetname{ToMi} benchmark when ablating the most selective ToM units and three sets of random units. We differentiate between results for questions that involves false-belief scenarios and true-belief ones. Our results indicate that we can localize specialized units for certain models, such as \modelname{Mistral-7B-Instruct}, but not for others, like \modelname{Phi-3.5-Mini-Instruct}. This differs from the findings related to the language network, where trends were consistent across all models (see Appendix \ref{app:finegrained-results}).

\section{Discussion}

\paragraph{Specialized LLM Language Units.}
Our findings provide compelling evidence that specialized language units emerge within LLMs. It is particularly surprising how effectively we can identify these units with the same limited set of localization stimuli employed in neuroscience, and that they prove to be causally relevant for language tasks.
While we observed consistent results across all 10 models we tested, it remains an open question whether this specialization is universal across all LLMs and under which conditions this specialization does or does not emerge. For instance, does the nature of the training data or the specific training objective influence the emergence of these specialized units? 
Moreover, the role of non-language-selective units remains unclear. It is possible they contribute to other specialized systems. While our experiments with the Multiple Demand and Theory of Mind selective units yielded some promising results, the variability across models suggests that these systems may either emerge more sparsely or be more complex or challenging to identify.

\paragraph{Consistency with the Brain's Language Network.}
Our paper adds to the growing body of research that uses neuroscience tools to interpret machine learning models \cite{schrimpf2020,schrimpf-pnas,zador2023catalyzing,tuckute2024language}. Specifically, our work takes a step towards analyzing LLM units that are \textit{causally} useful within a given system, providing a more stringent notion of functional correspondence \cite{cao2022putting,cao2024explanatory,mahon2023higher,princerepresentation}. 
The consistency between the causally important language units in LLMs and the human brain may suggest that computations, beyond representations, could be shared between these two systems.
This prompts an intriguing question: do specialized subsystems, such as the language network, always emerge as a consequence of optimizing for next-word prediction, and is such a simple objective the driver of specialization in the brain? 
Exploring this connection further could shed light on how cognitive processes evolve from such predictive mechanisms.

\paragraph{Related Work}
Previous work has identified a core language system within LLMs \cite{zhao2023unveiling}, but their approach requires finetuning the model on a next-token prediction task to locate parameters that exhibit minimal variation during finetuning. In contrast, our method bypasses additional training and leverages established research from language neuroscience. Concurrently, \citet{Sun2024BrainlikeFO} have shown that LLMs exhibit brain-like functional organization by using regressors to predict brain activity based on artificial neuron responses, and thereby mapping LLM representations onto the brain. However, their method is computationally intensive and lacks the cognitive neuroscience grounding that underpins our approach. Other efforts have focused on identifying subnetworks that encode factual knowledge \cite{meng2022locating, Bayazit2023DiscoveringKS, Hernandez2023InspectingAE} and task-specific skill neurons \cite{Panigrahi2023TaskSpecificSL}.

\paragraph{Future Work.}
Extending the analyses presented here to multimodal models could shed light on whether specialized Multiple Demand and Theory of Mind units are also responsive to non-linguistic inputs, regardless of the input modality (e.g., visual or auditory stimuli). This investigation aligns with the emergent modularity observed in the brain, where these networks are robustly dissociable from language \cite{mahowald2024dissociating}. In contrast, this dissociation is not evident in LLMs: ablating the language units renders the model incapable of understanding input sentences and, consequently, unable to perform any task presented verbally. This limitation applies to all tasks, as the input to LLMs is solely language-based.

\section{Conclusion}

In this paper, we explored whether functional specialization observed in the human brain can be identified in LLMs. Drawing inspiration from neuroscience, we applied the same localizers used in human neuroscience, to uncover language-selective units within LLMs, showing that a small subset of these units are crucial for language modeling. Our lesion studies revealed that ablating even a fraction of these units leads to significant drops in language performance across multiple benchmarks, while randomly sampled non-language units had no comparable effect. 
Although we successfully identified a language network analog in all models studied, we found mixed results when applying similar localization techniques to Theory of Mind and Multiple Demand networks, suggesting that not all cognitive functions neatly map onto current LLMs. 
These findings provide new insights into the internal structure of LLMs and open up avenues for further exploration of parallels between artificial systems and the human brain.


\section*{Limitations}


Our analysis focused on models smaller than 13 billion parameters, which may not capture the specialization that could emerge in larger models, such as those with 70 billion parameters. Additionally, we evaluated Theory of Mind (ToM) and Multiple Demand (MD) units using just one benchmark for each: \datasetname{ToMi} QA for ToM and a mathematical reasoning task (\datasetname{MATH}) for MD. While these benchmarks provided initial insights, they do not offer a comprehensive evaluation of these cognitive systems since our main focus was analyzing the language-selective units and their relationship to the human language network. Future work will involve expanding our study to include more models and a broader set of benchmarks to ensure robustness and generalizability. We also plan to analyze varying numbers of selective units for the MD and ToM networks, as this study focused only on the top 1\% which might not reflect the total number of units specialized for cognitively demanding tasks.

Moreover, the localizers we used to identify specialized units were adapted from neuroscience. While these methods allowed us to draw meaningful comparisons between LLMs and the brain, they are constrained by the stimuli sets traditionally used in neuroscience. Future work will consider developing more targeted and robust localizers that are not restricted by the same limitations, enabling deeper investigation into the specialization of LLMs across different tasks and domains.

\section*{Ethical Statement}

This research focuses on understanding the internal mechanisms of existing large language models (LLMs) by drawing parallels to human cognitive systems. Our work is aimed at advancing scientific knowledge in the field of AI and neuroscience and does not involve any human or animal subjects.

\bibliography{references}

\appendix

\clearpage


\section*{Appendix}

\section{Functional Localizers}
\label{app:functional-localizers}

Figure \ref{fig:localizer-stimuli} shows a pair of examples for each network localizer stimuli. We provide more details of each stimuli set below.

\subsection{Language Localizer}
\label{app:tom-localizer}

The language localizer uses the same set of 240 sentences and 240 lists of non-words\footnote{The language localizer stimuli were retrieved from: \url{https://www.evlab.mit.edu/resources-all/download-localizer-tasks}} as used by neuroscientists to localize the human language network. Each sentence consists of 12 words, and each list of non-words consists of 12 non-words to control for length. Since we are using a trained BPE tokenizer that breaks down each word into a number of tokens, we truncated the tokens to have a maximum length of 12 to ensure similar sequence length.

\subsection{Multiple Demand Localizer}
\label{app:md-localizer}

The arithmetic multiple-demand localizer used in neuroscience includes a set of ``hard'' arithmetic questions alongside a set of ``easy'' ones. These stimuli are usually generated by a MATLAB script that displays a mathematical problem on a screen for participants to solve, followed by two answer choices, one of which is correct. ``Hard'' questions are defined as addition problems with results exceeding 20 (e.g., 18+5), while ``easy'' questions yield results below 10 (e.g., 4+2). We adapted this localizer by similarly generating hard and easy arithmetic questions but slightly increased the complexity. Specifically, for hard questions, we sampled two numbers between 100 and 200, with each problem randomly chosen to be either an addition or subtraction with equal likelihood. For easy questions, we sampled two numbers in the range of 1 to 20. We generated 100 questions for each stimuli set. Examples are shown in Figure \ref{fig:localizer-stimuli}.

\subsection{Theory of Mind Localizer}
\label{app:tom-localizer}

\begin{figure*}[ht]
    \centering
    \includegraphics[width=1\linewidth]{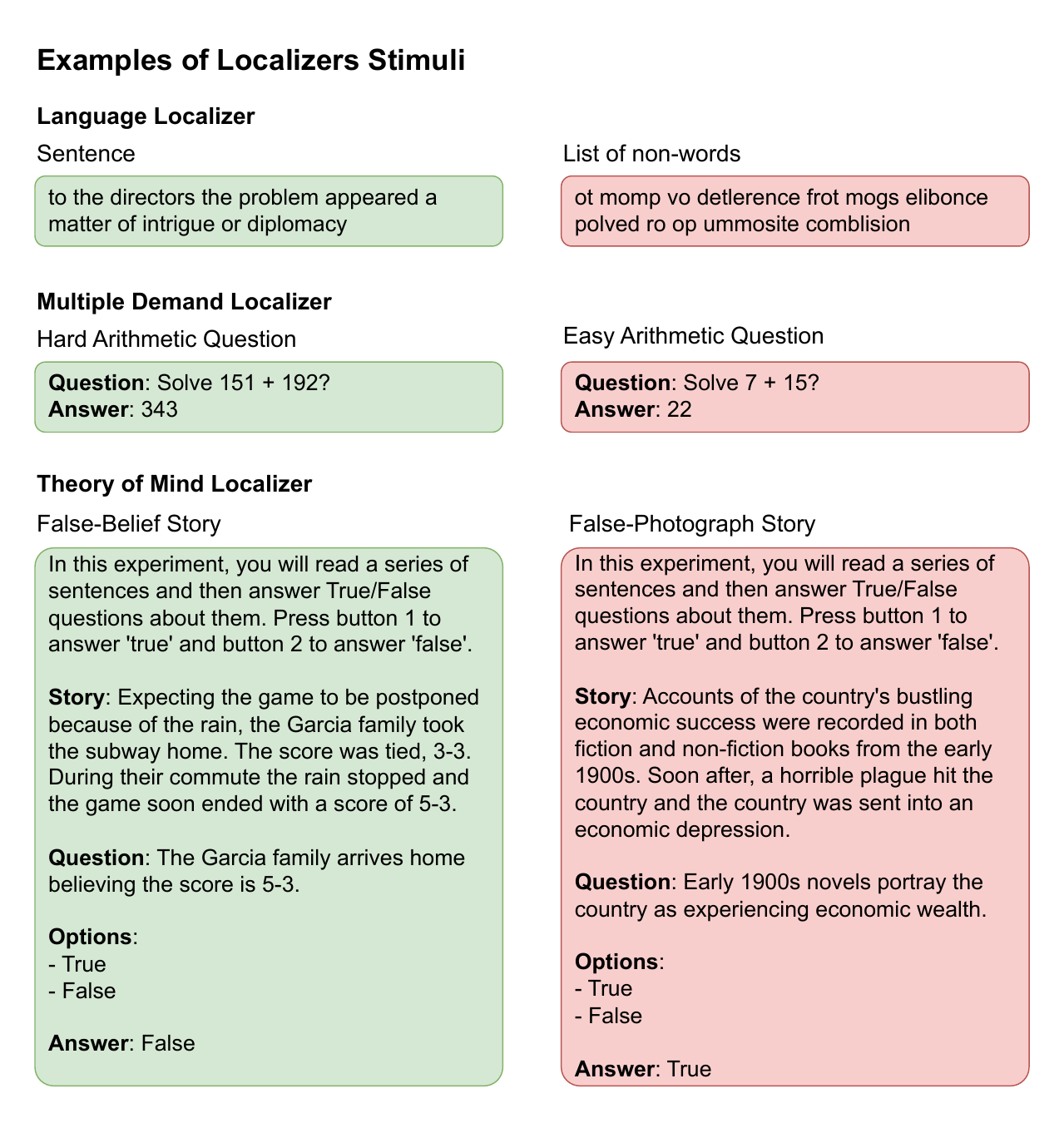}
    \caption{
        \textbf{Examples of Localizers Stimuli}.
        Language stimuli consists of 240 sentences and 240 lists of non-words.
        Multiple Demand stimuli consists of 100 hard arithmetic problems and 100 easy ones.
        Theory of Mind consists of 10 false-belief stories and 10 false-photograph stories.
        The instruction given in the Theory of Mind stimuli is the same given to participants during the neuroimaging study.
        See Appendix \ref{app:functional-localizers} for more details about each localizer.
    }
    \label{fig:localizer-stimuli}
\end{figure*}

We use the same set of stimuli employed in neuroscience to localize the theory-of-mind network in the human brain \cite{dodell2011tom}, which includes 10 false-belief stories and 10 false-photograph stories\footnote{The Theory of Mind localizer stimuli were retrieved from \url{https://saxelab.mit.edu/use-our-efficient-false-belief-localizer/}}. The prompt was structured to mirror the instructions given to participants during the neuroimaging study, followed by the story, the question, two answer choices (True or False), and an answer. Example of the prompt given from each set are shown in Figure \ref{fig:localizer-stimuli}. When evaluating the model on the test-set, we give it the following instruction: ``The following multiple choice questions is based on the following story. The question is related to Theory-of-Mind. Read the story and then answer the questions. Choose the best answer from the options provided by printing it as is without any modifications.''

\section{Localized Units Location}
\label{app:language-units-loc}

\subsection{Language Units}

\begin{figure*}
    \centering
    \includegraphics[width=1\linewidth]{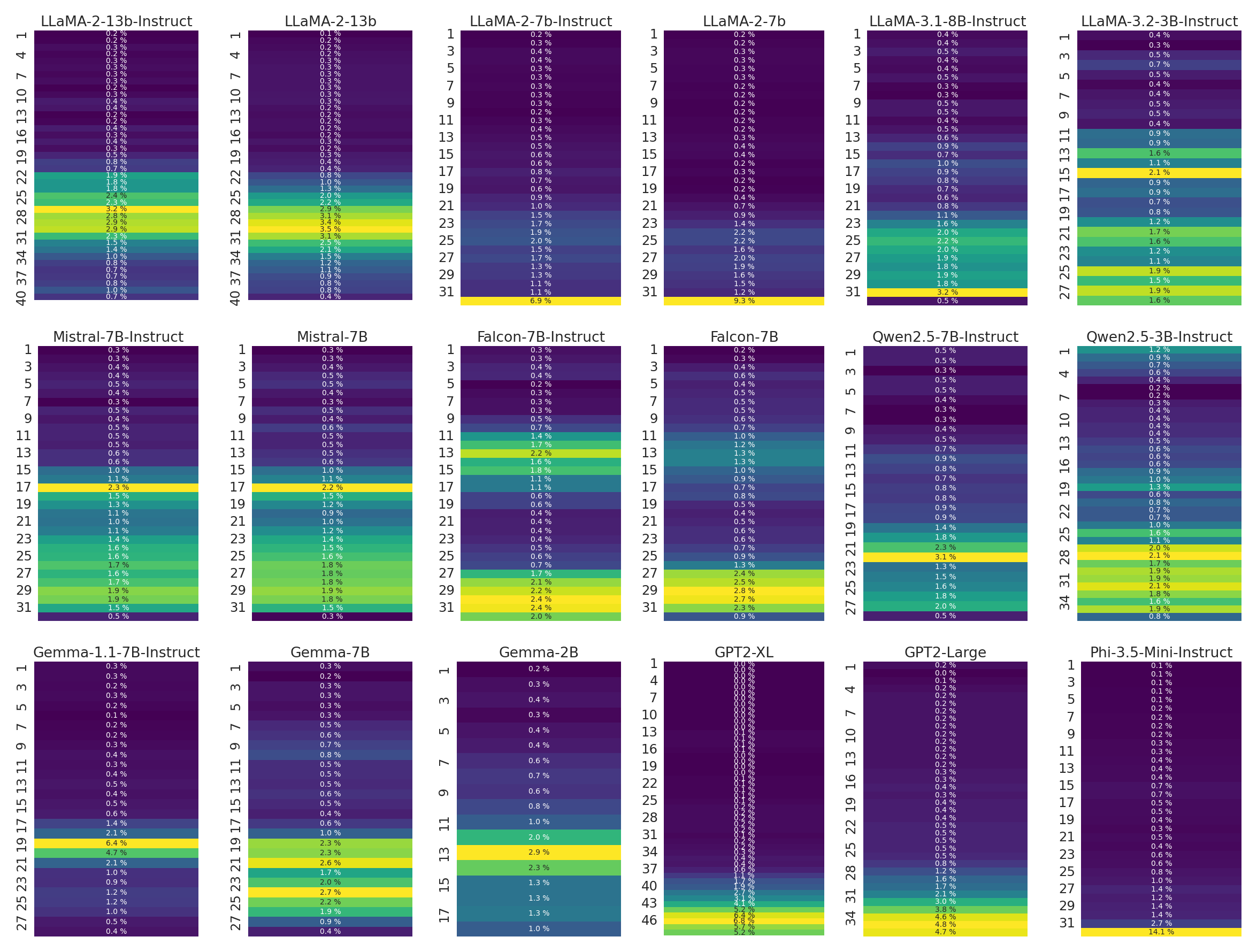}
    \caption{
        \textbf{Distribution of Language Units Across Layers}.
        The distribution of the top 1\% most language-selective units across layers in all 18 models tested in this work. The models are displayed from top to bottom, with each layer labeled by the percentage of units identified as belonging to the top 1\%.
    }
    \label{fig:lang-loc-app}
\end{figure*}

Figure \ref{fig:lang-loc-app} shows the distribution of the top 1\% language selective units for all 18 models tested in this work. An interesting observation is that the distribution of language-selective units remains similar in models both before and after instruction tuning.

\subsection{Multiple Demand Units}

\begin{figure*}
    \centering
    \includegraphics[width=1\linewidth]{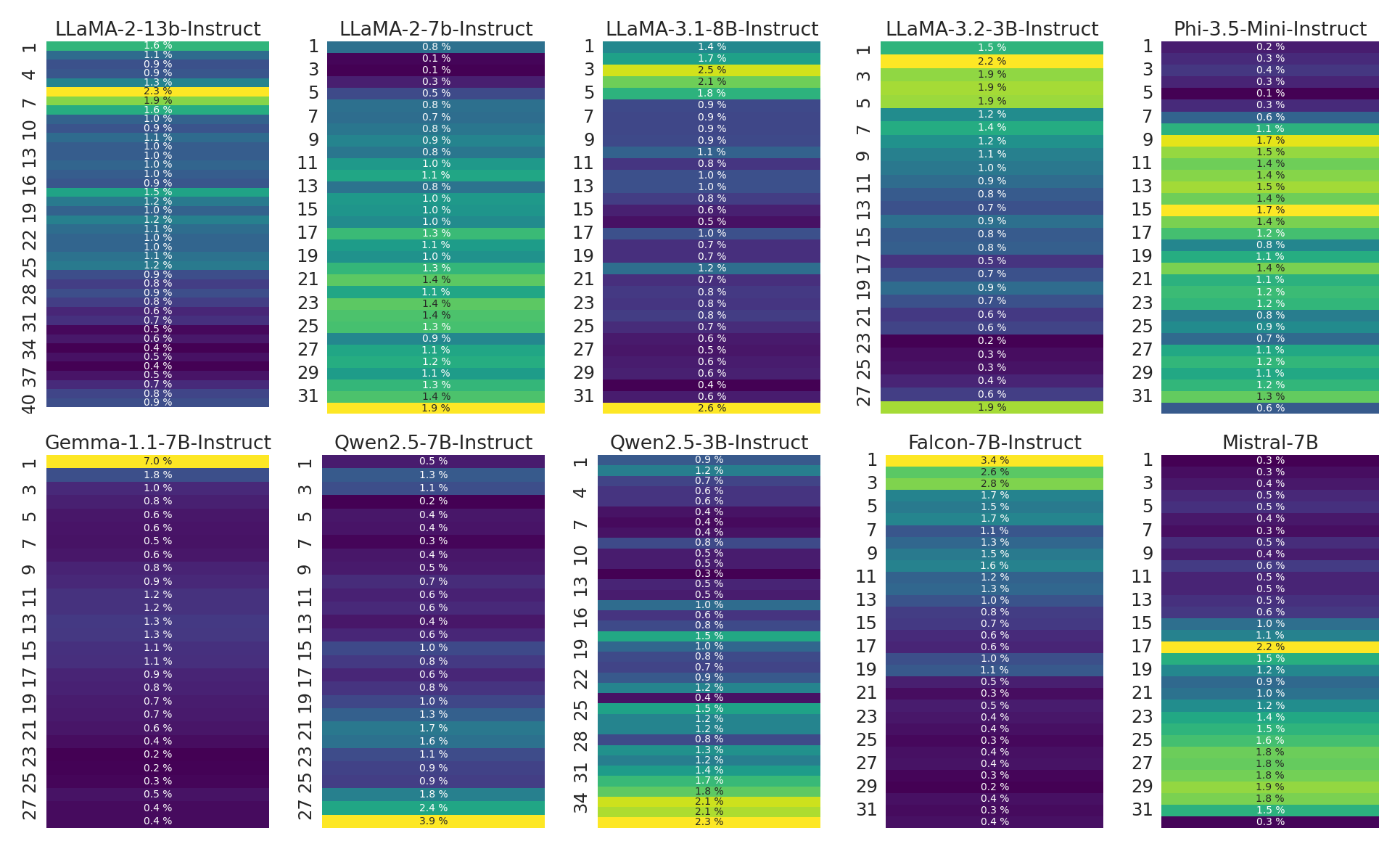}
    \caption{
        \textbf{Distribution of Multiple Demand Units Across Layers}.
        The distribution of the top 1\% most Multiple Demand (MD) selective units across layers in the 10 models tested for MD in this work. The models are displayed from top to bottom, with each layer labeled by the percentage of units identified as belonging to the top 1\%.
    }
    \label{fig:md-loc-app}
\end{figure*}

Figure \ref{fig:md-loc-app} shows the distribution of the top 1\% Multiple Demand (MD) selective units for the 10 models tested for MD in this work.

\subsection{Theory of Mind Units}

\begin{figure*}
    \centering
    \includegraphics[width=1\linewidth]{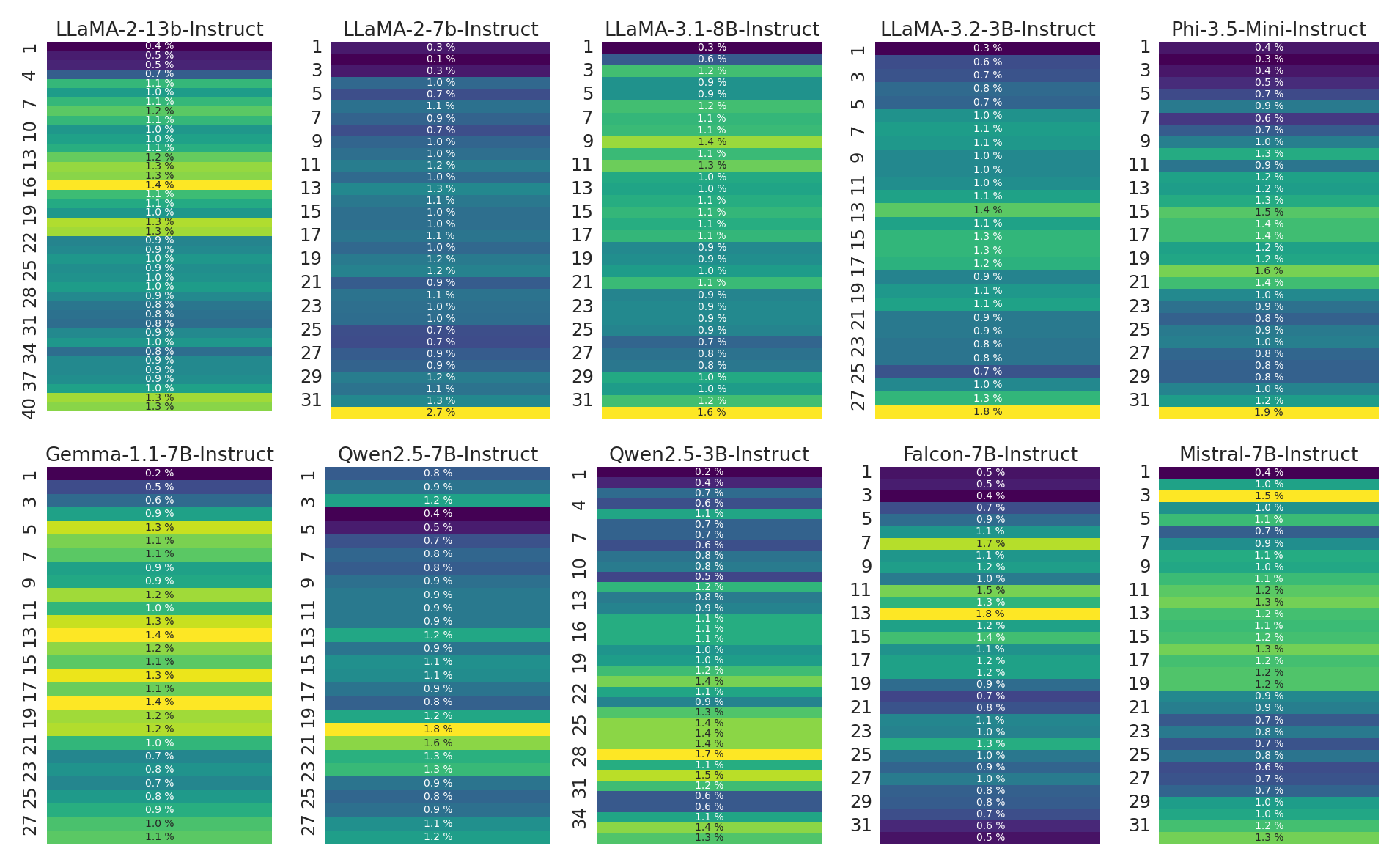}
    \caption{
        \textbf{Distribution of Theory of Mind Units Across Layers}.
        The distribution of the top 1\% most theory of mind (ToM) selective units across layers in the 10 models tested for ToM in this work. The models are displayed from top to bottom, with each layer labeled by the percentage of units identified as belonging to the top 1\%.
    }
    \label{fig:tom-loc-app}
\end{figure*}

Figure \ref{fig:tom-loc-app} shows the distribution of the top 1\% Theory of Mind (ToM) selective units for the 10 models tested for ToM in this work. The ToM selective units are more distributed across the model layers rather than being more clustered as in MD and the language-selective units. This might be due to the small number of stimuli samples used for the ToM localizer.

\section{Models}
\label{app:models}

Table \ref{tab:models} lists all 18 models analyzed in this study and indicates which models were used in which experiment. We kept 10 models for each experiment as shown in the last row. Table \ref{tab:num-units} shows the number of units corresponding to each percentage level for all models.

\begin{table}[ht]
\centering
\begin{tabular}{@{}llcc@{}}
\toprule
& \textbf{Model}           & \multicolumn{1}{l}{\textbf{Lang}} & \multicolumn{1}{l}{\textbf{MD/ToM}}  \\ \midrule
   & GPT2-Large               & \cmark                                    & \xmark           \\
   & GPT2-XL                  & \cmark                                    & \xmark            \\
   & Gemma-2B                 & \cmark                                    & \xmark             \\
   & Gemma-7B                 & \cmark                                    & \xmark             \\
   & Gemma-1.1-7B-Instruct    & \xmark                                    & \cmark              \\
   & Phi-3.5-Mini-Instruct    & \cmark                                    & \cmark              \\
   & Mistral-7B-v0.3          & \cmark                                    & \xmark             \\
   & Mistral-7B-Instruct-v0.3 & \xmark                                    & \cmark              \\
   & LLaMA-2-7B               & \cmark                                    & \xmark             \\
   & LLaMA-2-7B-Chat          & \xmark                                    & \cmark              \\
   & LLaMA-2-13B              & \cmark                                    & \xmark             \\
   & LLaMA-2-13B-Chat         & \xmark                                    & \cmark              \\       
   & LLaMA-3.1-8B-Instruct    & \cmark                                    & \cmark              \\
   & LLaMA-3.2-3B-Instruct    & \xmark                                    & \cmark              \\
   & Falcon-7B                & \cmark                                    & \xmark             \\
   & Falcon-7B-Instruct       & \xmark                                    & \cmark              \\
   & Qwen2.5-3B-Instruct      & \xmark                                    & \cmark              \\
   & Qwen2.5-7B-Instruct      & \xmark                                    & \cmark              \\ \midrule
 \textbf{\#}         & 18   & 10                                    & 10                      
\\ \bottomrule
\end{tabular}
\caption{
  \textbf{Models Used in This Work}
  Overview of the 18 models analyzed, with an indication of the specific experiments in which each model was used. Lang referes to the language experiments, MD refers to the Multiple Demand experiments, and ToM refers to the Theory of Mind experiments.
}
\label{tab:models}
\end{table}

\section{Finegrained Results}
\label{app:finegrained-results}

\subsection{Language Results}
Tables \ref{tab:lang-results} and \ref{tab:lang-results-2} display results for the 10 models tested on three language benchmarks---SyntaxGym, BLiMP, and GLUE---along with the average performance across these benchmarks. Each model's performance is shown initially without ablation, followed by ablations of the top-{0.125, 0.25, 0.5, 1}\% language-selective units, and then with randomly sampled units at the same percentages. The performance changes in Figure \ref{fig:lesion-study} can be reproduced by subtracting post-ablation results from the baseline (0\%) for both language-selective and random unit ablations. Results with random units are averaged across three random seeds.

\subsection{Multiple Demand Results}
Table \ref{tab:md-results} presents the results for the 10 models tested on the \datasetname{MATH} benchmark, organized by difficulty level and including the overall macro average across levels. Each model’s performance is shown under three conditions: without ablation, after ablating the top 1\% of Multiple Demand-selective units, and with an equivalent number of randomly sampled units.

\subsection{Theory of Mind Results}
Table \ref{tab:tom-results} similarly shows the results for the 10 models tested on the \datasetname{ToMi} benchmark, organized by question type, whether it involves a false-belief scenario (n=231) or true-belief scenarios (n=389), and including the macro average across both types. Each model’s performance is shown under three conditions: without ablation, after ablating the top 1\% of theory-of-mind-selective units, and with an equivalent number of randomly sampled units. Table \ref{tab:tom-results-2} shows the same but when ablating the top-2\% of units.

\section{More Brain Alignment Statistical Tests}
\label{app:stats-test}

In Section \ref{sec:llm-brain-similarity}, we performed Welch's t-test to demonstrate that units from the language network are significantly more brain-aligned than three sets of randomly sampled units from the model, particularly when sampling a small number of units. Here, we conduct the Shapiro-Wilk test to verify that each distribution follows a normal distribution, as Welch’s t-test assumes normality in the compared distributions. Tables \ref{tab:shapiro-wilk-pereira2018} and \ref{tab:shapiro-wilk-tuckute2024} present the test statistics and p-values for the brain alignment results across models, comparing both language and random units at each percentage and for each dataset. These results confirm that the distributions are indeed normal. A p-value greater than 0.05 indicates normality, while values below this threshold suggest deviation from normality. Notably, the only cases where the p-value falls below 0.05—indicating non-normal distributions—are for the 0.5\% and 1\% conditions in the Tuckute2024 dataset, where no significant difference was observed.

\begin{table}[h]
    \centering
    \begin{tabular}{l r r}
        \toprule
        Percentage & Language Units  & Random Units \\
        \midrule
        0.03125\% & (0.902, 0.233) & (0.971, 0.565) \\
        0.0625\%  & (0.957, 0.755) & (0.939, 0.085) \\
        0.125\%   & (0.955, 0.722) & (0.981, 0.853) \\
        0.25\%    & (0.933, 0.475) & (0.962, 0.345) \\
        0.5\%     & (0.945, 0.609) & (0.974, 0.658) \\
        1\%       & (0.945, 0.612) & (0.970, 0.551) \\
        \bottomrule
    \end{tabular}
    \caption{Shapiro-Wilk test (statistics and p-values) for brain alignment distributions across models. The test is conducted separately for language units and randomly sampled units at each percentage level for the \datasetname{Pereira2018} dataset.}
    \label{tab:shapiro-wilk-pereira2018}
\end{table}

\begin{table}[h]
    \centering
    \begin{tabular}{l r r}
        \toprule
        Percentage & Language Units & Random Units \\
        \midrule
        0.03125\% & (0.948, 0.641) & (0.9511, 0.181) \\
        0.0625\%  & (0.962, 0.802) & (0.9507, 0.176) \\
        0.125\%   & (0.973, 0.915) & (0.9810, 0.852) \\
        0.25\%    & (0.914, 0.309) & (0.9592, 0.296) \\
        0.5\%     & (0.829, 0.032) & (0.9693, 0.519) \\
        1\%       & (0.825, 0.029) & (0.9725, 0.608) \\
        \bottomrule
    \end{tabular}
    \caption{Shapiro-Wilk test (statistics and p-values) for brain alignment distributions across models. The test is conducted separately for language units and randomly sampled units at each percentage level for the \datasetname{Tuckute2024} dataset.}
    \label{tab:shapiro-wilk-tuckute2024}
\end{table}

We also conducted a permutation test, a non-parametric statistical method that does not require the assumption of normality. This method involves randomly shuffling data labels across 10,000 permutations to generate a null distribution of the test statistic. By comparing the observed test statistic to this null distribution, we evaluated the statistical significance of our results. The findings from the permutation test confirmed the significance of our results, as shown in Table \ref{tab:permutation-test}.

\begin{table}[h]
    \centering
    \begin{tabular}{l r r}
        \toprule
        Percentage & \datasetname{Pereira2018} & \datasetname{Tuckute2024} \\
        \midrule
        0.03125\%     & 0.001  & 0.004  \\
        0.0625\%     & 0.004  & 0.013  \\
        0.125\%    & 0.138  & 0.000  \\
        0.25\%     & 0.548  & 0.013  \\
        0.5\%     & 0.195  & 0.169  \\
        1\%     & 0.084  & 0.458  \\
        \bottomrule
    \end{tabular}
    \caption{Permutation test p-values assessing the statistical significance of brain alignment differences on both datasets. The test was conducted by randomly shuffling data labels across 10,000 permutations to generate a null distribution of the test statistic. The observed test statistic was then compared to this null distribution to compute the p-values. Lower p-values indicate stronger evidence against the null hypothesis, confirming the robustness of our findings.}
    \label{tab:permutation-test}
\end{table}

\section{Brain-Score Datasets}
\label{app:brain-alignment}

\paragraph{Tuckute2024}
This dataset consists of 5 participants reading 1000 6-word sentences presented one sentence at a time for 2s. BOLD responses from voxels in the language network were averaged within each participant and then across participants to yield an overall average language network response to each sentence. The stimuli used span a large part of the linguistic space, enabling model-brain comparisons across a wide range of single sentences. Sentence presentation order was randomized across participants. The averaging of sentences across participants effectively minimizes the effect of temporal autocorrelation/context in this dataset. In combination with the diversity in linguistic materials, this dataset presents a particularly challenging dataset for model evaluation. The noise ceiling for \datasetname{Tuckute2024} is $r = 0.56$,  see \citet{tuckute2024driving} for more details.

\paragraph{Pereira2018}
This dataset consists of fMRI activations (blood-oxygen-level-dependent; BOLD responses) recorded as participants read short passages presented one sentence at a time for 4 s. The dataset is composed of two distinct experiments: one with 9 subjects presented with 384 sentences, and another with 6 subjects presented with 243 sentences each. The passages in each experiment spanned 24 different topics. The results reported for this dataset are the average alignment across both experiments \cite{pereira_toward_2018}. The reported results for this dataset use an unshuffled cross-validation scheme, ensuring that sentences from the same passage appear exclusively in either the training or testing set.

\begingroup

\renewcommand{\arraystretch}{0.9} 

\begin{table*}[]
\centering
\begin{tabular}{@{}lllccc|c@{}}
\toprule
\textbf{Model} & \textbf{Ablation Units} & \textbf{Percentage} & \textbf{SyntaxGym} & \textbf{BLiMP} & \textbf{GLUE} & \textbf{Average} \\ \midrule
\multirow{9}{*}{GPT2-Large} & - & 0\% & 78.50 & 83.55 & 45.42 &  69.16 \\ \cmidrule{2-7} 
& Language & 0.125\% & 61.87 & 76.79 & 44.23 &  60.96 \\ 
& Language & 0.25\% & 50.03 & 72.69 & 43.13 &  55.28 \\ 
& Language & 0.5\% & 46.99 & 69.37 & 38.55 &  51.64 \\ 
& Language & 1\% & 41.07 & 66.09 & 39.96 &  49.04 \\ 
\cmidrule{2-7} 
& Random & 0.125\% & 78.02 & 83.50 & 45.39 &  68.97 \\ 
& Random & 0.25\% & 78.28 & 83.33 & 45.49 &  69.03 \\ 
& Random & 0.5\% & 77.95 & 82.89 & 44.48 &  68.44 \\ 
& Random & 1\% & 76.87 & 82.70 & 43.97 &  67.85 \\ 
\midrule 
\multirow{9}{*}{GPT2-XL} & - & 0\% & 82.70 & 83.38 & 46.85 &  70.98 \\ \cmidrule{2-7} 
& Language & 0.125\% & 80.20 & 81.64 & 44.78 &  68.88 \\ 
& Language & 0.25\% & 70.73 & 78.06 & 46.24 &  65.01 \\ 
& Language & 0.5\% & 66.54 & 77.19 & 44.75 &  62.82 \\ 
& Language & 1\% & 56.02 & 74.86 & 43.12 &  58.00 \\ 
\cmidrule{2-7} 
& Random & 0.125\% & 82.26 & 83.16 & 46.40 &  70.61 \\ 
& Random & 0.25\% & 80.76 & 83.07 & 46.38 &  70.07 \\ 
& Random & 0.5\% & 79.93 & 82.68 & 45.53 &  69.38 \\ 
& Random & 1\% & 79.44 & 81.64 & 45.19 &  68.76 \\ 
\midrule 
\multirow{9}{*}{Gemma-2B} & - & 0\% & 80.15 & 81.14 & 47.81 &  69.70 \\ \cmidrule{2-7} 
& Language & 0.125\% & 38.16 & 56.34 & 41.79 &  45.43 \\ 
& Language & 0.25\% & 36.59 & 54.52 & 39.82 &  43.64 \\ 
& Language & 0.5\% & 26.02 & 52.54 & 37.38 &  38.64 \\ 
& Language & 1\% & 25.46 & 51.60 & 37.56 &  38.21 \\ 
\cmidrule{2-7} 
& Random & 0.125\% & 80.18 & 81.10 & 47.35 &  69.54 \\ 
& Random & 0.25\% & 79.49 & 80.88 & 48.42 &  69.60 \\ 
& Random & 0.5\% & 79.51 & 80.93 & 46.25 &  68.90 \\ 
& Random & 1\% & 65.89 & 72.20 & 42.65 &  60.25 \\ 
\midrule 
\multirow{9}{*}{Gemma-7B} & - & 0\% & 80.37 & 81.75 & 62.29 &  74.80 \\ \cmidrule{2-7} 
& Language & 0.125\% & 54.99 & 64.30 & 43.34 &  54.21 \\ 
& Language & 0.25\% & 52.91 & 61.17 & 44.38 &  52.82 \\ 
& Language & 0.5\% & 25.67 & 63.75 & 41.15 &  43.52 \\ 
& Language & 1\% & 29.61 & 48.97 & 45.90 &  41.50 \\ 
\cmidrule{2-7} 
& Random & 0.125\% & 80.15 & 80.48 & 61.96 &  74.20 \\ 
& Random & 0.25\% & 80.44 & 81.24 & 60.59 &  74.09 \\ 
& Random & 0.5\% & 80.55 & 81.25 & 63.05 &  74.95 \\ 
& Random & 1\% & 79.65 & 79.98 & 58.36 &  72.66 \\ 
\midrule 
\multirow{9}{*}{Phi-3.5-Mini-Instruct} & - & 0\% & 81.86 & 80.63 & 70.73 &  77.74 \\ \cmidrule{2-7} 
& Language & 0.125\% & 45.42 & 58.62 & 60.60 &  54.88 \\ 
& Language & 0.25\% & 34.81 & 55.72 & 49.65 &  46.72 \\ 
& Language & 0.5\% & 25.37 & 53.56 & 33.40 &  37.44 \\ 
& Language & 1\% & 22.90 & 53.79 & 46.26 &  40.98 \\ 
\cmidrule{2-7} 
& Random & 0.125\% & 80.16 & 80.95 & 70.80 &  77.30 \\ 
& Random & 0.25\% & 81.83 & 81.64 & 69.64 &  77.70 \\ 
& Random & 0.5\% & 78.79 & 80.35 & 68.61 &  75.92 \\ 
& Random & 1\% & 79.80 & 79.05 & 69.19 &  76.01 \\   \bottomrule
\end{tabular}
\caption{
   \textbf{Language Benchmarks Results 1}
   Results for the 5 models on the language benchmarks tested in this work. Random for each percentage is averaged across 3 seeds. The results when ablating random units is almost the same as ablating no units, while ablating language units lead to a sharp drop in performance. See Table \ref{tab:lang-results-2} for the results of the other models.
}
\label{tab:lang-results}
\end{table*}

\endgroup

\begingroup

\renewcommand{\arraystretch}{0.9} 

\begin{table*}[]
\centering
\begin{tabular}{@{}lllccc|c@{}}
\toprule
\textbf{Model} & \textbf{Ablation Units} & \textbf{Percentage} & \textbf{SyntaxGym} & \textbf{BLiMP} & \textbf{GLUE} & \textbf{Average} \\ \midrule
\multirow{9}{*}{LLaMA-2-7b} & - & 0\% & 81.07 & 85.63 & 50.60 &  72.43 \\ \cmidrule{2-7} 
& Language & 0.125\% & 46.07 & 66.85 & 41.91 &  51.61 \\ 
& Language & 0.25\% & 39.51 & 64.24 & 40.91 &  48.22 \\ 
& Language & 0.5\% & 28.86 & 57.07 & 32.57 &  39.50 \\ 
& Language & 1\% & 26.82 & 56.01 & 38.33 &  40.39 \\ 
\cmidrule{2-7} 
& Random & 0.125\% & 81.09 & 85.57 & 50.74 &  72.47 \\ 
& Random & 0.25\% & 81.26 & 85.03 & 50.25 &  72.18 \\ 
& Random & 0.5\% & 80.23 & 84.68 & 51.15 &  72.02 \\ 
& Random & 1\% & 80.63 & 84.53 & 47.44 &  70.87 \\ 
\midrule 
\multirow{9}{*}{LLaMA-2-13b} & - & 0\% & 82.91 & 84.82 & 59.53 &  75.76 \\ \cmidrule{2-7} 
& Language & 0.125\% & 78.57 & 81.38 & 48.05 &  69.33 \\ 
& Language & 0.25\% & 62.12 & 74.84 & 42.47 &  59.81 \\ 
& Language & 0.5\% & 23.85 & 51.23 & 29.16 &  34.75 \\ 
& Language & 1\% & 29.13 & 51.42 & 30.03 &  36.86 \\ 
\cmidrule{2-7} 
& Random & 0.125\% & 82.43 & 84.79 & 58.76 &  75.33 \\ 
& Random & 0.25\% & 82.13 & 84.66 & 55.18 &  73.99 \\ 
& Random & 0.5\% & 82.06 & 83.77 & 57.52 &  74.45 \\ 
& Random & 1\% & 81.21 & 83.55 & 53.94 &  72.90 \\ 
\midrule 
\multirow{9}{*}{LLaMA-3.1-8B-Instruct} & - & 0\% & 80.05 & 81.90 & 69.20 &  77.05 \\ \cmidrule{2-7} 
& Language & 0.125\% & 80.25 & 79.60 & 66.44 &  75.43 \\ 
& Language & 0.25\% & 78.22 & 76.96 & 61.43 &  72.20 \\ 
& Language & 0.5\% & 73.12 & 77.60 & 55.77 &  68.83 \\ 
& Language & 1\% & 54.12 & 67.17 & 46.36 &  55.88 \\ 
\cmidrule{2-7} 
& Random & 0.125\% & 79.93 & 81.89 & 68.98 &  76.93 \\ 
& Random & 0.25\% & 79.99 & 81.88 & 68.71 &  76.86 \\ 
& Random & 0.5\% & 79.92 & 81.10 & 69.51 &  76.85 \\ 
& Random & 1\% & 79.14 & 81.73 & 67.41 &  76.09 \\ 
\midrule 
\multirow{9}{*}{Mistral-7B} & - & 0\% & 80.39 & 83.44 & 64.03 &  75.95 \\ \cmidrule{2-7} 
& Language & 0.125\% & 70.08 & 75.38 & 47.33 &  64.26 \\ 
& Language & 0.25\% & 44.11 & 66.73 & 44.91 &  51.91 \\ 
& Language & 0.5\% & 37.60 & 66.39 & 43.74 &  49.24 \\ 
& Language & 1\% & 33.05 & 61.85 & 40.34 &  45.08 \\ 
\cmidrule{2-7} 
& Random & 0.125\% & 80.28 & 83.34 & 63.54 &  75.72 \\ 
& Random & 0.25\% & 80.46 & 83.13 & 63.91 &  75.84 \\ 
& Random & 0.5\% & 80.34 & 82.62 & 62.75 &  75.24 \\ 
& Random & 1\% & 79.22 & 82.51 & 63.00 &  74.91 \\ 
\midrule 
\multirow{9}{*}{Falcon-7B} & - & 0\% & 80.05 & 80.35 & 48.36 &  69.59 \\ \cmidrule{2-7} 
& Language & 0.125\% & 72.17 & 75.83 & 46.86 &  64.95 \\ 
& Language & 0.25\% & 69.99 & 71.67 & 47.23 &  62.97 \\ 
& Language & 0.5\% & 51.36 & 60.23 & 44.17 &  51.92 \\ 
& Language & 1\% & 25.79 & 55.42 & 45.11 &  42.10 \\ 
\cmidrule{2-7} 
& Random & 0.125\% & 79.59 & 80.26 & 48.44 &  69.43 \\ 
& Random & 0.25\% & 79.89 & 80.35 & 48.83 &  69.69 \\ 
& Random & 0.5\% & 78.62 & 79.96 & 48.40 &  69.00 \\ 
& Random & 1\% & 78.32 & 79.99 & 48.85 &  69.05 \\   \bottomrule
\end{tabular}
\caption{
   \textbf{Language Benchmarks Results 2}
   Results for 5 models on the language benchmarks tested in this work. Random for each percentage is averaged across 3 seeds. The results when ablating random units is almost the same as ablating no units, while ablating language units lead to a sharp drop in performance. See Table \ref{tab:lang-results} for the results of the other models.
}
\label{tab:lang-results-2}
\end{table*}

\endgroup

\begingroup

\renewcommand{\arraystretch}{0.9} 

\begin{table*}[]
\centering
\begin{tabular}{@{}llccccc|c@{}}
\toprule
\textbf{Model} & \textbf{Ablation}  & \textbf{Level 1} & \textbf{Level 2} & \textbf{Level 3} & \textbf{Level 4} & \textbf{Level 5} & \textbf{Average} \\ \midrule 
\multirow{3}{*}{Phi-3.5-Mini-Instruct} & - & 52.33 & 41.50 & 40.45 & 35.11 & 31.91 &  40.26 \\ \cmidrule{2-8} 
& MD & 38.37 & 32.31 & 33.90 & 28.44 & 27.33 &  32.07 \\ 
& Random & 49.92 & 43.50 & 41.52 & 37.31 & 35.71 &  41.59 \\ 
\midrule 
\multirow{3}{*}{LLaMA-3.1-8B-Instruct} & - & 40.00 & 37.41 & 36.23 & 36.36 & 39.21 &  37.84 \\ \cmidrule{2-8} 
& MD & 37.21 & 34.13 & 33.18 & 33.61 & 40.45 &  35.72 \\ 
& Random & 36.20 & 35.15 & 33.96 & 35.95 & 38.59 &  35.97 \\ 
\midrule 
\multirow{3}{*}{Mistral-7B-Instruct} & - & 39.07 & 35.71 & 37.31 & 35.61 & 34.01 &  36.34 \\ \cmidrule{2-8} 
& MD & 36.28 & 33.90 & 32.65 & 32.03 & 30.51 &  33.07 \\ 
& Random & 35.35 & 33.07 & 34.05 & 34.03 & 33.23 &  33.95 \\ 
\midrule 
\multirow{3}{*}{LLaMA-2-7b-Instruct} & - & 24.42 & 28.57 & 29.24 & 28.69 & 29.04 &  27.99 \\ \cmidrule{2-8} 
& MD & 24.65 & 29.71 & 29.42 & 29.44 & 29.11 &  28.47 \\ 
& Random & 23.41 & 28.46 & 27.50 & 26.94 & 28.39 &  26.94 \\ 
\midrule 
\multirow{3}{*}{LLaMA-2-13b-Instruct} & - & 25.35 & 33.11 & 29.78 & 29.19 & 29.35 &  29.35 \\ \cmidrule{2-8} 
& MD & 28.14 & 25.74 & 24.48 & 25.27 & 23.84 &  25.49 \\ 
& Random & 26.51 & 31.18 & 28.34 & 28.86 & 28.73 &  28.72 \\ 
\midrule 
\multirow{3}{*}{LLaMA-3.2-3B-Instruct} & - & 35.35 & 32.77 & 33.99 & 33.69 & 35.56 &  34.27 \\ \cmidrule{2-8} 
& MD & 31.63 & 32.20 & 33.90 & 33.11 & 34.70 &  33.11 \\ 
& Random & 34.19 & 32.77 & 32.56 & 34.50 & 35.74 &  33.95 \\ 
\midrule 
\multirow{3}{*}{Gemma-1.1-7B-Instruct} & - & 40.00 & 37.41 & 35.96 & 34.28 & 35.87 &  36.71 \\ \cmidrule{2-8} 
& MD & 35.58 & 34.81 & 30.49 & 31.28 & 32.53 &  32.94 \\ 
& Random & 37.21 & 36.28 & 35.19 & 33.75 & 35.12 &  35.51 \\ 
\midrule 
\multirow{3}{*}{Falcon-7B-Instruct} & - & 23.49 & 26.64 & 23.86 & 25.85 & 23.91 &  24.75 \\ \cmidrule{2-8} 
& MD & 27.91 & 26.30 & 25.20 & 25.19 & 24.15 &  25.75 \\ 
& Random & 26.98 & 25.21 & 24.48 & 25.05 & 23.96 &  25.14 \\ 
\midrule 
\multirow{3}{*}{Qwen2.5-7B-Instruct} & - & 59.53 & 60.43 & 59.37 & 56.46 & 56.91 &  58.54 \\ \cmidrule{2-8} 
& MD & 59.07 & 56.12 & 56.59 & 55.55 & 54.66 &  56.40 \\ 
& Random & 57.75 & 57.60 & 56.53 & 54.05 & 53.73 &  55.93 \\ 
\midrule 
\multirow{3}{*}{Qwen2.5-3B-Instruct} & - & 53.49 & 49.55 & 49.33 & 44.37 & 47.28 &  48.80 \\ \cmidrule{2-8} 
& MD & 43.72 & 40.14 & 40.54 & 36.03 & 38.66 &  39.82 \\ 
& Random & 42.95 & 40.36 & 39.13 & 36.86 & 37.50 &  39.36 \\ 
\bottomrule
\end{tabular}
\caption{
   \textbf{MATH Benchmark Results}
   Results for the 10 models tested on the MATH benchmark, showing performance in the following conditions for each model: without ablation, with ablation of the top 1\% most MD-selective, and with randomly sampled. The results for \emph{Random} is averaged across 3 seeds. MD stands for multiple demand.
}
\label{tab:md-results}
\end{table*}

\endgroup

\begingroup

\renewcommand{\arraystretch}{0.9} 

\begin{table*}[]
\centering
\begin{tabular}{@{}llcc|c@{}}
\toprule
\textbf{Model} & \textbf{Ablation Units}  & \textbf{False Belief} & \textbf{True Belief} & \textbf{Average} \\ \midrule 
\multirow{3}{*}{Phi-3.5-Mini-Instruct} & - & 50.65 & 86.38 &  68.51 \\ \cmidrule{2-5} 
& ToM & 17.75 & 98.46 &  58.10 \\ 
& Random & 17.75 & 96.92 &  57.33 \\ 
\midrule 
\multirow{3}{*}{LLaMA-3.1-8B-Instruct} & - & 80.95 & 75.32 &  78.14 \\ \cmidrule{2-5} 
& ToM & 64.50 & 75.32 &  69.91 \\ 
& Random & 76.62 & 68.47 &  72.54 \\ 
\midrule 
\multirow{3}{*}{Mistral-7B-Instruct} & - & 79.22 & 65.81 &  72.52 \\ \cmidrule{2-5} 
& ToM & 64.07 & 61.95 &  63.01 \\ 
& Random & 69.55 & 68.47 &  69.01 \\ 
\midrule 
\multirow{3}{*}{LLaMA-2-7b-Instruct} & - & 23.81 & 79.69 &  51.75 \\ \cmidrule{2-5} 
& ToM & 20.78 & 79.95 &  50.36 \\ 
& Random & 32.47 & 68.38 &  50.42 \\ 
\midrule 
\multirow{3}{*}{LLaMA-2-13b-Instruct} & - & 63.64 & 68.38 &  66.01 \\ \cmidrule{2-5} 
& ToM & 49.35 & 69.92 &  59.64 \\ 
& Random & 59.88 & 58.01 &  58.95 \\ 
\midrule 
\multirow{3}{*}{LLaMA-3.2-3B-Instruct} & - & 9.96 & 92.80 &  51.38 \\ \cmidrule{2-5} 
& ToM & 9.52 & 91.77 &  50.65 \\ 
& Random & 16.88 & 85.52 &  51.20 \\ 
\midrule 
\multirow{3}{*}{Gemma-1.1-7B-Instruct} & - & 78.79 & 65.55 &  72.17 \\ \cmidrule{2-5} 
& ToM & 62.77 & 66.07 &  64.42 \\ 
& Random & 72.87 & 63.58 &  68.23 \\ 
\midrule 
\multirow{3}{*}{Falcon-7B-Instruct} & - & 50.22 & 46.02 &  48.12 \\ \cmidrule{2-5} 
& ToM & 49.78 & 45.50 &  47.64 \\ 
& Random & 52.67 & 48.41 &  50.54 \\ 
\midrule 
\multirow{3}{*}{Qwen2.5-7B-Instruct} & - & 97.84 & 41.65 &  69.74 \\ \cmidrule{2-5} 
& ToM & 93.51 & 44.73 &  69.12 \\ 
& Random & 92.35 & 46.62 &  69.48 \\ 
\midrule 
\multirow{3}{*}{Qwen2.5-3B-Instruct} & - & 81.82 & 59.38 &  70.60 \\ \cmidrule{2-5} 
& ToM & 77.06 & 56.56 &  66.81 \\ 
& Random & 46.90 & 60.07 &  53.48 \\ 
\bottomrule
\end{tabular}
\caption{
   \textbf{TOMi Benchmark Results (1\% Lesion)}
   Results for the 10 models tested on the TOMi benchmark, showing performance in the following conditions for each model: without ablation, with ablation of the top 1\% most ToM-selective, and with randomly sampled. The results for \emph{Random} is averaged across 3 seeds. ToM stands for theory of mind.
}
\label{tab:tom-results}
\end{table*}

\endgroup

\begingroup

\renewcommand{\arraystretch}{0.9} 

\begin{table*}[]
\centering
\begin{tabular}{@{}llcc|c@{}}
\toprule
\textbf{Model} & \textbf{Ablation Units}  & \textbf{False Belief} & \textbf{True Belief} & \textbf{Average} \\ \midrule 
\multirow{3}{*}{Phi-3.5-Mini-Instruct} & - & 50.65 & 86.38 &  68.51 \\ \cmidrule{2-5} 
& ToM & 7.36 & 97.69 &  52.52 \\ 
& Random & 27.71 & 92.03 &  59.87 \\ 
\midrule 
\multirow{3}{*}{LLaMA-3.1-8B-Instruct} & - & 80.95 & 75.32 &  78.14 \\ \cmidrule{2-5} 
& ToM & 49.35 & 64.52 &  56.94 \\ 
& Random & 61.62 & 56.56 &  59.09 \\ 
\midrule 
\multirow{3}{*}{Mistral-7B-Instruct} & - & 79.22 & 65.81 &  72.52 \\ \cmidrule{2-5} 
& ToM & 40.26 & 71.98 &  56.12 \\ 
& Random & 66.67 & 66.50 &  66.58 \\ 
\midrule 
\multirow{3}{*}{LLaMA-2-7b-Instruct} & - & 23.81 & 79.69 &  51.75 \\ \cmidrule{2-5} 
& ToM & 19.05 & 77.63 &  48.34 \\ 
& Random & 35.93 & 63.75 &  49.84 \\ 
\midrule 
\multirow{3}{*}{LLaMA-2-13b-Instruct} & - & 63.64 & 68.38 &  66.01 \\ \cmidrule{2-5} 
& ToM & 42.42 & 60.15 &  51.29 \\ 
& Random & 52.53 & 54.07 &  53.30 \\ 
\midrule 
\multirow{3}{*}{LLaMA-3.2-3B-Instruct} & - & 9.96 & 92.80 &  51.38 \\ \cmidrule{2-5} 
& ToM & 19.48 & 82.01 &  50.74 \\ 
& Random & 25.83 & 76.86 &  51.35 \\ 
\midrule 
\multirow{3}{*}{Gemma-1.1-7B-Instruct} & - & 78.79 & 65.55 &  72.17 \\ \cmidrule{2-5} 
& ToM & 61.04 & 67.87 &  64.45 \\ 
& Random & 71.57 & 64.27 &  67.92 \\ 
\midrule 
\multirow{3}{*}{Falcon-7B-Instruct} & - & 50.22 & 46.02 &  48.12 \\ \cmidrule{2-5} 
& ToM & 52.81 & 46.27 &  49.54 \\ 
& Random & 50.07 & 50.64 &  50.36 \\ 
\midrule 
\multirow{3}{*}{Qwen2.5-7B-Instruct} & - & 97.84 & 41.65 &  69.74 \\ \cmidrule{2-5} 
& ToM & 89.61 & 50.13 &  69.87 \\ 
& Random & 60.03 & 59.13 &  59.58 \\ 
\midrule 
\multirow{3}{*}{Qwen2.5-3B-Instruct} & - & 81.82 & 59.38 &  70.60 \\ \cmidrule{2-5} 
& ToM & 87.45 & 37.28 &  62.36 \\ 
& Random & 64.07 & 46.02 &  55.04 \\ 
\bottomrule
\end{tabular}
\caption{
   \textbf{TOMi Benchmark Results (2\% Lesion)}
   Results for the 10 models tested on the TOMi benchmark, showing performance in the following conditions for each model: without ablation, with ablation of the top 2\% most ToM-selective, and with randomly sampled. The results for \emph{Random} is averaged across 3 seeds. ToM stands for theory of mind.
}
\label{tab:tom-results-2}
\end{table*}

\endgroup

\begin{table*}[ht]
\centering
\begin{tabular}{@{}lccccccc@{}}
\toprule
\textbf{Model}  &  \textbf{0.03125\%} &  \textbf{0.0625\%} &  \textbf{0.125\%} &  \textbf{0.25\%} &  \textbf{0.5\%} &  \textbf{1\%} &  \textbf{2\%} \\
\midrule
\textbf{Falcon-7B}             &       45 &       90 &      181 &      363 &      727 &     1454 &     2908 \\
\textbf{Falcon-7B-Instruct}    &       45 &       90 &      181 &      363 &      727 &     1454 &     2908 \\
\textbf{GPT2-Large}            &       14 &       28 &       57 &      115 &      230 &      460 &      921 \\
\textbf{GPT2-XL}               &       24 &       48 &       96 &      192 &      384 &      768 &     1536 \\
\textbf{Gemma-1.1-7B-Instruct} &       26 &       53 &      107 &      215 &      430 &      860 &     1720 \\
\textbf{Gemma-2B}              &       11 &       23 &       46 &       92 &      184 &      368 &      737 \\
\textbf{Gemma-7B}              &       26 &       53 &      107 &      215 &      430 &      860 &     1720 \\
\textbf{LLaMA-2-13b}           &       64 &      128 &      256 &      512 &     1024 &     2048 &     4096 \\
\textbf{LLaMA-2-13b-Instruct}  &       64 &      128 &      256 &      512 &     1024 &     2048 &     4096 \\
\textbf{LLaMA-2-7b}            &       40 &       81 &      163 &      327 &      655 &     1310 &     2621 \\
\textbf{LLaMA-2-7b-Instruct}   &       40 &       81 &      163 &      327 &      655 &     1310 &     2621 \\
\textbf{LLaMA-3.1-8B-Instruct} &       40 &       81 &      163 &      327 &      655 &     1310 &     2621 \\
\textbf{LLaMA-3.2-3B-Instruct} &       26 &       53 &      107 &      215 &      430 &      860 &     1720 \\
\textbf{Mistral-7B}            &       40 &       81 &      163 &      327 &      655 &     1310 &     2621 \\
\textbf{Mistral-7B-Instruct}   &       40 &       81 &      163 &      327 &      655 &     1310 &     2621 \\
\textbf{Phi-3.5-Mini-Instruct} &       30 &       61 &      122 &      245 &      491 &      983 &     1966 \\
\textbf{Qwen2.5-3B-Instruct}   &       23 &       46 &       92 &      184 &      368 &      737 &     1474 \\
\textbf{Qwen2.5-7B-Instruct}   &       31 &       62 &      125 &      250 &      501 &     1003 &     2007 \\
\bottomrule
\end{tabular}
\caption{
  \textbf{Number of Units at Specified Percentage Levels for Each Model}
  The table shows the number of units corresponding to each percentage level (x\%) for each model. These values are calculated by multiplying the number of layers, by the hidden dimension, and the specified percentage.
}
\label{tab:num-units}
\end{table*}

\end{document}